\documentclass[review]{elsarticle}

\usepackage{lineno,hyperref}
\modulolinenumbers[5]

\journal{Neural Networks}

\usepackage{amsmath}
\usepackage{multirow}
\usepackage{booktabs}

\usepackage{amsmath}

\begin{document}

\begin{frontmatter}

\title{Emphasizing Unseen Words: New Vocabulary Acquisition for End-to-End Speech Recognition}

\author[mymainaddress,mysecaddress]{Leyuan Qu\corref{mycorrespondingauthor}}
\cortext[mycorrespondingauthor]{Corresponding author}

\ead[url]{http://www.informatik.uni-hamburg.de/WTM/}

\ead{quleyuan9826@gmail.com}

\author[mymainaddress]{Cornelius Weber}
\ead{cornelius.weber@uni-hamburg.de}

\author[mymainaddress]{Stefan Wermter}
\ead{stefan.wermter@uni-hamburg.de}


\address[mymainaddress]{Knowledge Technology, Department of Informatics, University of Hamburg, Hamburg, Germany}
\address[mysecaddress]{Department of Artificial Intelligence, Zhejiang Lab, Hangzhou, China}

\begin{abstract}
Due to the dynamic nature of human language, automatic speech recognition (ASR) systems need to continuously acquire new vocabulary.
Out-Of-Vocabulary (OOV) words, such as trending words and new named entities, pose problems to modern ASR systems that require long training times to adapt their large numbers of parameters.
Different from most previous research focusing on language model post-processing, we tackle this problem on an earlier processing level and eliminate the bias in acoustic modeling to recognize OOV words acoustically. We propose to generate OOV words using text-to-speech systems and to rescale losses to encourage neural networks to pay more attention to OOV words. Specifically, we enlarge the classification loss used for training neural networks' parameters of utterances containing OOV words (sentence-level), or rescale the gradient used for back-propagation for OOV words (word-level), when fine-tuning a previously trained model on synthetic audio. To overcome catastrophic forgetting, we also explore the combination of loss rescaling and model regularization, i.e.\ L2 regularization and elastic weight consolidation (EWC). Compared with previous methods that just fine-tune synthetic audio with EWC, the experimental results on the LibriSpeech benchmark reveal that our proposed loss rescaling approach can achieve significant improvement on the recall rate with only a slight decrease on word error rate. Moreover, word-level rescaling is more stable than utterance-level rescaling and leads to higher recall rates and precision on OOV word recognition. Furthermore, our proposed combined loss rescaling and weight consolidation methods can support continual learning of an ASR system.

\end{abstract}


\end{frontmatter}


\section{Introduction}

Recently, end-to-end ASR models have been receiving a lot of attention and achieving impressive performance~\cite{graves2006connectionist, graves2012sequence, bahdanau2014neural}. These models significantly simplify the training process to directly map acoustic inputs to characters or words. Additionally, limited domain-specific knowledge is required, which dramatically boosts model development and deployment. However, end-to-end models need a lot of training data and perform poorly on words out-of-vocabulary (OOV) or rarely existing in the training data, for example, trending words and new named entities. 

Since it takes substantial efforts to collect labeled OOV speech data for ASR model training, current approaches to tackle the OOV problem mainly involve a language model (LM) or post-processing, for instance, user-dependent language models~\cite{brown1992class,maskey2004improved}, LM rescoring~\cite{guo2019spelling} and finite-state transducer lattice extension~\cite{zhao2019shallow}. However, the post-processing techniques only obtain limited improvement as they do not tackle the root causes at the acoustic level. 

Alternatively, fine-tuning end-to-end ASR models with synthetic audio containing OOV words can efficiently improve the recall rate of unseen vocabulary, which usually leverages advanced text-to-speech (TTS) systems to generate audio-text pairs required for ASR model training. However, the catastrophic forgetting problem substantially degrades the overall performance of ASR systems, especially on non-OOV words. Elastic weight consolidation (EWC)~\cite{kirkpatrick2017overcoming} is adapted to tackle this problem but leads to a limited recall rate improvement for OOV word recognition. In this paper, we take this method a step further and propose loss rescaling to encourage models to pay more attention to unknown words. Instead of just fine-tuning ASR models where all words are treated equally, enlarging the loss of utterances containing OOV words (sentence-level) or increasing the gradient of unseen words (word-level) can efficiently incline the model to update the weights related to OOV words. We choose 100 OOV words appearing in LRS3-TED dataset but not existing in LibriSpeech dataset. Then, we crawl texts including the new words from the Internet and synthesize audio with TTS systems. The experimental results of fine-tuning audio-text pairs on a hybrid CTC\footnote{CTC is the abbreviation for Connectionist Temporal Classification.}/attention ASR model show a significant improvement on recall. When combining EWC with the word-level loss rescaling, we achieve 45.81\% of recall on the ROOV test with only 7.8\% and 4.6\% of relative WER increase on the LibriSpeech test-clean and test-other data sets respectively. As a result, we have improved the recognition of OOV words while maintaining the accuracy for non-OOV words.

\section{Related Work}

\subsection{The Recognition of OOV Words in End-to-End ASR Models}
\label{section: oov word review}

Since the OOV problem has occurred, a number of approaches have been proposed for conventional GMM-HMM\footnote{GMM and HMM are short for Gaussian Mixture Model and Hidden Markov Model respectively.} models~\cite{sheikh2017modelling, miao2013deep} and hybrid DNN-HMM\footnote{DNN is short for Deep Neural Network.} models~\cite{knill2013investigation,hwang2016character}. In this section, we only review methods towards end-to-end ASR architectures which have been the most promising methods in speech recognition in recent years.

Aleksic et al.~\cite{aleksic2015improved} extend class-based LMs~\cite{brown1992class,maskey2004improved} by creating a user-dependent small LM for contact name recognition on voice commands, which is compiled dynamically based on the contact names on users' devices. Moreover, contact insertion reward is proposed to avoid excessive bias and to balance the information between user-dependent and user-independent cases. Hori et al.~\cite{hori2017multi} combine word-level with character-level language modeling in end-to-end architectures. With the word-level LM, the model can achieve better performance by learning stronger and longer context information, while character-level LM is used to overcome the OOV issue that the word-level LM suffers from. A similar idea is investigated by Li et al.\cite{li2017acoustic} on acoustic-to-word mapping employing character-level modeling units to tackle OOV issues. Williams et al.~\cite{williams2018contextual} leverage contextual information, for instance, users' locations, users' favorite songs, and calendar events, to partially rescore the output likelihood from sequence-to-sequence models during beam search instead of bringing an additional LM in. Since previous work does not consider errors generated by speech recognition systems when combining them with external LMs, Guo et al.~\cite{guo2019spelling} incorporate a spelling correction model into the speech recognizer training, that directly maps speech recognizer outputs to ground-truth texts. The experimental results suggest that the proposed spelling correction model outperforms n-best LM rescoring and TTS data fine-tuning. 

To enable on-device end-to-end speech recognition models to individually recognize new named entities, such as the contact names on mobile phones, Sim et al.~\cite{sim2019personalization} compare LM biasing and acoustic model fine-tuning methods. Furthermore, several techniques, such as layer freezing, early stopping, and EWC, are investigated to suppress model overfitting during fine-tuning. Instead of using a word-level LM in ASR in which a pre-defined lexicon is required,  Likhomanenko et al.~\cite{likhomanenko2019needs} attempt to decode acoustic models with a character-level LM which is not constrained by lexicons. The lexicon-free decoder achieves better results on OOV experiments since the character-level LM is naturally able to handle unseen words. Different from traditional hybrid ASR models, for instance DNN-HMM, end-to-end ASR architectures aim to learn LMs and acoustic models into one module, which leads to no clear division between language and acoustic models in the end-to-end fashion. To combine LMs trained from a different domain, Variani et al.~\cite{variani2020hybrid} propose hybrid autoregressive transducer (HAT) to separately model internal LM used for training and an external LM from a different domain used for inference. Consequently, Meng et al.~\cite{meng2021internal} improve HAT by estimating and subtracting the internal LM scores, and properly integrate an external LM into end-to-end ASR architectures without any further training. To further improve the model performance on proper nouns, Zhao et al.~\cite{zhao2019shallow} optimize the shallow-fusion method~\cite{kannan2018analysis} (integrate an external LM into the inference of a sequence-to-sequence model) by building the LM at subword level instead of at word level. Additionally, they propose the early contextual finite state transducer (FST) to avoid the proper noun candidates being pruned during the Viterbi beam search. Moreover, a common set of prefixes is utilized to avoid the contextual biasing always being active and prevent models from degrading on cases not containing OOV words.

Different from most of the previous work focusing on LM post-processing which requires candidate units existing in n-best lists or decoding lattices, in this paper, we tackle the OOV problem on an earlier processing level by eliminating the bias in acoustic modeling to recognize OOV words acoustically.

\subsection{Data Augmentation with Synthetic Audio for ASR}
\label{section: tts for asr}
 Proper speech data augmentation does not only boost the model performance, but can also significantly improve system robustness and generalization~\cite{ko2015audio}. There are many strategies used in ASR training, for example, noise addition, pitch shifting, speed perturbation, back-translation~\cite{hayashi2018back} and room impulse response injection with real or simulated data~\cite{ko2017study}. More recently, a simple yet effective approach, SpecAugment~\cite{park2019specaugment}, has been proposed and achieves state-of-the-art results on the LibriSpeech benchmark corpus. The basic idea for SpecAugment is randomly masking or cropping a fixed area on spectrograms in the time- or frequency-domain, which effectively prevents model overfitting, especially for noisy conditions. Another well-established method is mixing synthetic audio with real data by leveraging advanced TTS models, like Tacotron2~\cite{shen2018natural}, DeepVoice3~\cite{ping2018deep} and FlowTron~\cite{valle2020flowtron}.
 
 Rossenbach et al.~\cite{rossenbach2020generating} compare commonly used data augmentation strategies with the TTS audio. The results reveal the effectiveness of TTS data in ASR system training. Laptev et al.~\cite{laptev2020you} investigate the effect of augmenting data with TTS audio for low-resource speech recognition. The resulting models outperform other systems with the same setting and semi-supervised learning methods. Furthermore, other authors explore the influence of the audio quality with different vocoders, i.e.\ Griffin-Lim and LPCNet~\cite{valin2019lpcnet}. Instead of just mixing the synthetic audio data during training, Rosenberg et al.~\cite{rosenberg2019speech} exploit the impact of the TTS model's effectiveness and diversity on ASR results. Moreover, lexical diversity is also investigated on domain adaptation experiments.
 
 Inspired by the benefit brought by TTS data, we synthesize audio with text crawled from the Internet containing OOV words as the training set for new vocabulary acquisition and model adaptation.
 
\subsection{End-to-End ASR Architectures}
\label{end-to-end review}
End-to-end ASR architectures aim to directly map acoustic observations to text transcripts, such as characters and words. Different from conventional systems which separately train the acoustic model and LM with different criteria and corpora, end-to-end ASR systems train all modules jointly. Consequently, the training progress is significantly simplified and does not require too much domain-specific knowledge. There are four main categories in end-to-end learning, i.e.\ CTC, Recurrent Neural Network Transducer (RNN-T), encoder-decoder with attention and hybrid CTC/attention architectures. They mainly differ in their alignment of input acoustic features and output label sequences.

 \subsubsection{CTC-based End-to-End Models}
 
 The basic idea of CTC-based approaches is to bring in a special token, 'blank', which is dynamically filled in the places between modeling units. Consequently, the exact boundary information required by conventional methods is not needed anymore. Then, a carefully designed dynamic programming algorithm is used to search optimal paths and convert the frame-level token sequences to meaningful utterances by removing blank tokens and merging repeated labels.
 
 Graves et al.~\cite{graves2006connectionist} firstly adopt the ground-breaking CTC approach to overcome the problems faced in conventional ASR systems, i.e.\ the requirement of frame-level segmentation and the mapping from model outputs to ground-truth labels. Hannun et al.~\cite{hannun2014first} simplify the CTC building process and propose a modified prefix-search decoding algorithm to completely discard the cumbersome decoding strategies used in HMM-based systems. Amodei et al.~\cite{amodei2016deep} conduct a comprehensive evaluation on model architecture, system optimization, and model deployment. Furthermore, parallel training on model level~\cite{coates2013deep} and data level~\cite{dean2012large} is utilized to heavily speed up the training process.
 
 Furthermore, substituting RNNs with Convolutional Neural Networks (CNNs) has been a growing trend since pure CNN-based architectures can dramatically reduce the training time and inference latency, which is critical for speech recognition tasks running in real time. Jasper~\cite{li2019jasper} achieves state-of-the-art results on the LibriSpeech dataset by stacking the CNN-only blocks. Inspired by Jasper, Krimany et al.~\cite{kriman2020quartznet} propose QuartzNet which uses a similar fully  convolutional architecture but with significantly fewer parameters. QuartzNet enables the CTC-based end-to-end ASR model to run locally on mobile devices.

\subsubsection{RNN-T End-to-End Models}

CTC assumes that every output node is conditionally independent of other outputs, which makes it difficult to model the dependency between adjacent frames. Consequently, RNN-T~\cite{graves2012sequence} was proposed to overcome the unreasonable conditionally independence assumptions. Different from CTC which only focuses on acoustic sequence modeling, RNN-T uses a separate module, called prediction network, to model the context information, which can be treated as an intrinsic language model, followed by a joint network to classify the concatenation of encoder and prediction network outputs.
 
 Rao et al.~\cite{rao2017exploring} exploit some strategies for RNN-T architecture training and find that CTC-based encoder pre-training and language model-based prediction network pre-training improve the model performance. Zhang et al.~\cite{zhang2020transformer} present a Transformer-based~\cite{vaswani2017attention} RNN-T model, in which the Transformer modules are used to learn representations from speech signals and text sequences. Inspired by Jasper~\cite{li2019jasper} and QuartzNet~\cite{kriman2020quartznet}, Han et al.~\cite{han2020contextnet} propose ContextNet by introducing a fully convolutional encoder into the RNN-T architecture. Since the CNN has a weaker receptivity for long context than Long Short-Term Memory (LSTM) and Transformers, the squeeze-and-excitation (SE) layer is integrated into ContextNet to enhance long-distance dependence. Gulati et al.~\cite{gulati2020conformer} propose Conformer which combines CNNs with self-attention~\cite{vaswani2017attention} to concurrently learn local and global features, and show significant improvement on the LibriSpeech test and test-other set.

\subsubsection{Attention-based Encoder-Decoder Models}

Another branch of end-to-end systems is the attention-based encoder-decoder architecture. Different from CTC or RNN-T architectures, the attention mechanism dynamically aligns the encoder and decoder time steps to temporally align the input and output sequences. Chorowski et al.~\cite{chorowski2015attention} transfer the attention-based recurrent networks to speech recognition and achieve comparable results on the TIMIT benchmark that compare well with conventional methods. Chan et al.~\cite{chan2015listen} propose LAS (listen, attend and spell) to directly map acoustic inputs to transcription. The results show that the attention mechanism prevents model overfitting on the training set. Meanwhile, Bahdanau et al.~\cite{bahdanau2016end} show that attention-based models can implicitly learn better context information than CTC and conventional models. In addition, local monotonic attention~\cite{tjandra2017local}, full-sequence attention~\cite{prabhavalkar2017analysis}, time-restricted self-attention~\cite{povey2018time}, multi-channel attention~\cite{braun2018multi} and online attention~\cite{fan2018online} are proposed to reduce the complexity of attention computation and learn more robust alignments.

\subsubsection{Hybrid CTC/Attention Architectures}

To fully incorporate the merits of CTC and attention models, Kim et al.~\cite{kim2017joint} propose to jointly train CTC and attention-based approaches in a multi-task learning fashion by sharing one encoder. The evaluation on WSJ and CHiME-4 noisy speech shows the hybrid architecture can efficiently speed up convergence and learn more robust alignment between input frames and output sequences. Hori et al.~\cite{hori2017joint} extend the hybrid CTC/attention method with a joint decoding algorithm by rescoring or combining the probabilities from both objective functions. Then, a monotonic chunk-wise attention~\cite{miao2019online} and transformer-based encoder~\cite{miao2020transformer} are utilized to enable the hybrid CTC/attention model to work in online streaming tasks. Zhang et al.~\cite{zhang2020unified} propose a new two-pass approach (U2) which unifies the streaming and non-streaming ASR models into one architecture. The hybrid CTC/attention architecture is becoming more and more popular. There has been a trend to unify the streaming and non-streaming architecture into one model with this architecture. However, limited research has focused on the recognition of OOV words in end-to-end architectures. In this paper, we tackle the problem of OOV words with the U2 ASR model\footnote{https://github.com/wenet-e2e/wenet} which will be described in more detail in ~\ref{sec:u2}.

\section{Methodology}
In this section, we demonstrate the proposed loss rescaling approaches at sentence level and word level. Furthermore, we introduce the techniques of L2 regularization and EWC used to overcome catastrophic forgetting problems.
 
\subsection{Loss Rescaling at Sentence Level}

\begin{figure}[!t]
  \centering
  \includegraphics[width=11cm]{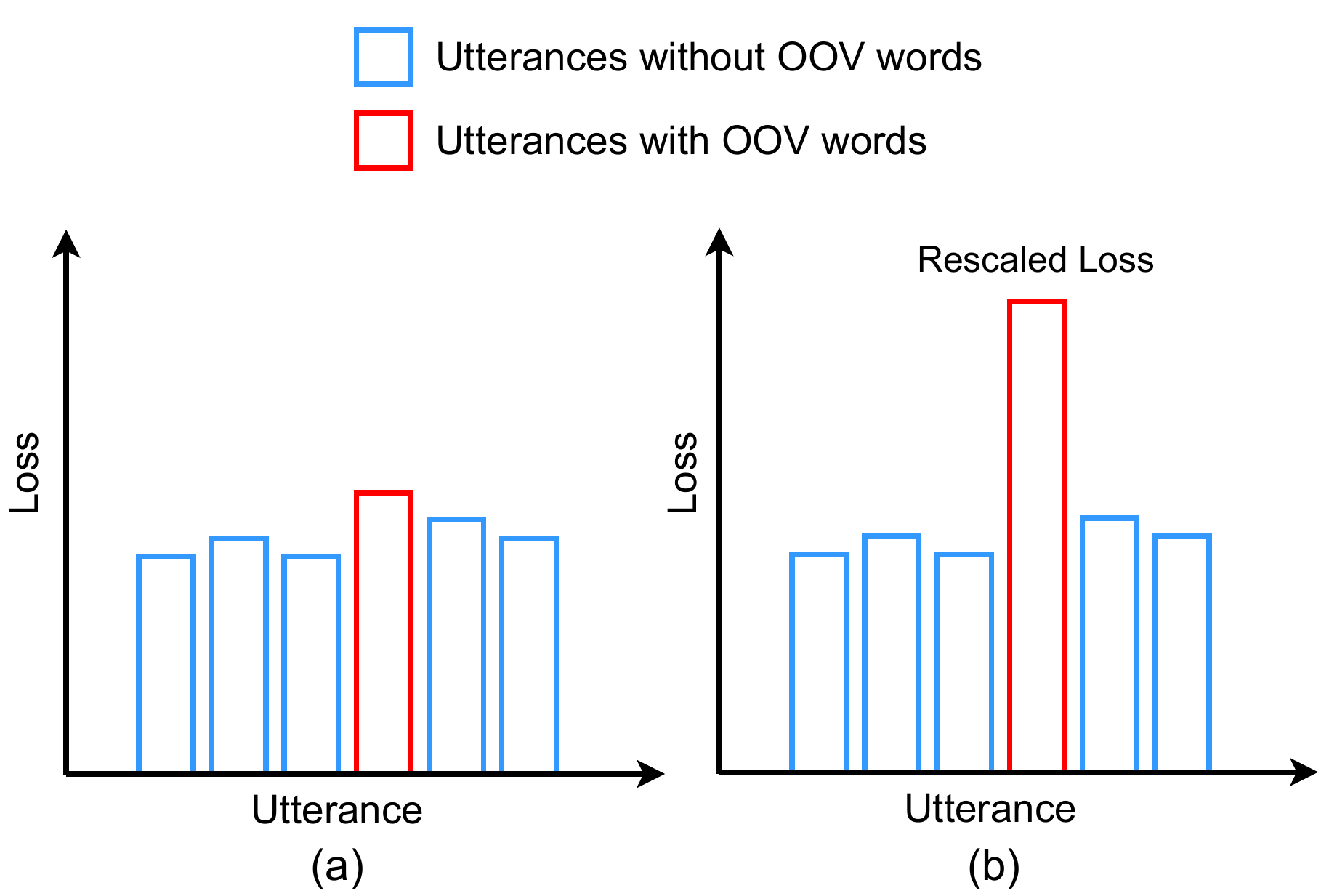}  
  \caption{(a) Utterance loss distribution in one mini-batch. (b) Utterance loss distribution after loss rescaling.}
  \label{fig:loss-distribution}
\end{figure}

During training, the CTC function returns one loss per utterance, and the mean of all utterance losses in the same mini-batch would be used for back-propagation. As shown in Figure~\ref{fig:loss-distribution} (a), each bar in the figure means one utterance loss in a randomly selected mini-batch. We observe that the utterance losses in one mini-batch are evenly distributed. Sometimes, the loss of utterances containing OOV words can be slightly higher or even lower than other utterances without OOV words. Consequently, the model pays equal attention to each utterance or word, which leads to the final model performance heavily relying on the frequency of words in training sets. 
Sometimes, the model attention is even biased by non-OOV words.

To emphasize an utterance containing OOV words, we rescale the utterance loss,
as indicated in Figure~\ref{fig:loss-distribution} (b),
by multiplying it with a hyper-parameter $\mu$ in Eq.~\eqref{sentence level}, where $x$ are acoustic inputs, $y$ are the target text references, and $O$ is the set of OOV words.

\begin{equation}
\label{sentence level}
\mathcal{L}_{sentence}(x,y)=
\begin{cases}
\mathcal{L}_{CTC}(x,y), &\text{if } o \text{ not in } y, \forall o \in O,\\
\mu \mathcal{L}_{CTC}(x,y), &\text{if } o \text{ in } y, \forall o \in O
\end{cases}
\end{equation}

\subsection{Loss Rescaling at Word Level}

Given an input acoustic vector $\textbf{x} = (x_{0},\cdot\cdot\cdot,x_{T})$, and a target label sequence $\textbf{y} = (y_{0},\cdot\cdot\cdot,y_{U})$, where $T>>U$, and $T$ and $U$ are the length of the acoustic vector and target label sequence respectively, the CTC loss aims to maximize the log probability in Eq.~\eqref{word1}, where $\widetilde{\textbf{y}}$ is the extended label sequence of $\textbf{y}$ by inserting blank labels $\phi$ at the beginning and the end of $\textbf{y}$ and between every two label tokens, $\mathcal{\widetilde{Y}}=\mathcal{Y}\cup\phi$. When training CTC ASR systems, the last layer of the neural network outputs one $N\times1$ vector for each acoustic frame $x_{t}$ at time step $t$, where $N$ is the number of modeling units. For example, when building on character level, $N$ is 28 ($a ... z $ + space + blank).

\begin{equation}
\label{word1}
\mathcal{L}_{CTC}=-P(\widetilde{\textbf{y}}|\textbf{x})=-\sum_{\textbf{a}\in\mathcal{F}^{-1}(\textbf{y})}P(\textbf{a}|\textbf{x})
\end{equation}

\begin{figure}[!t]
\centering
\includegraphics[width=12cm]{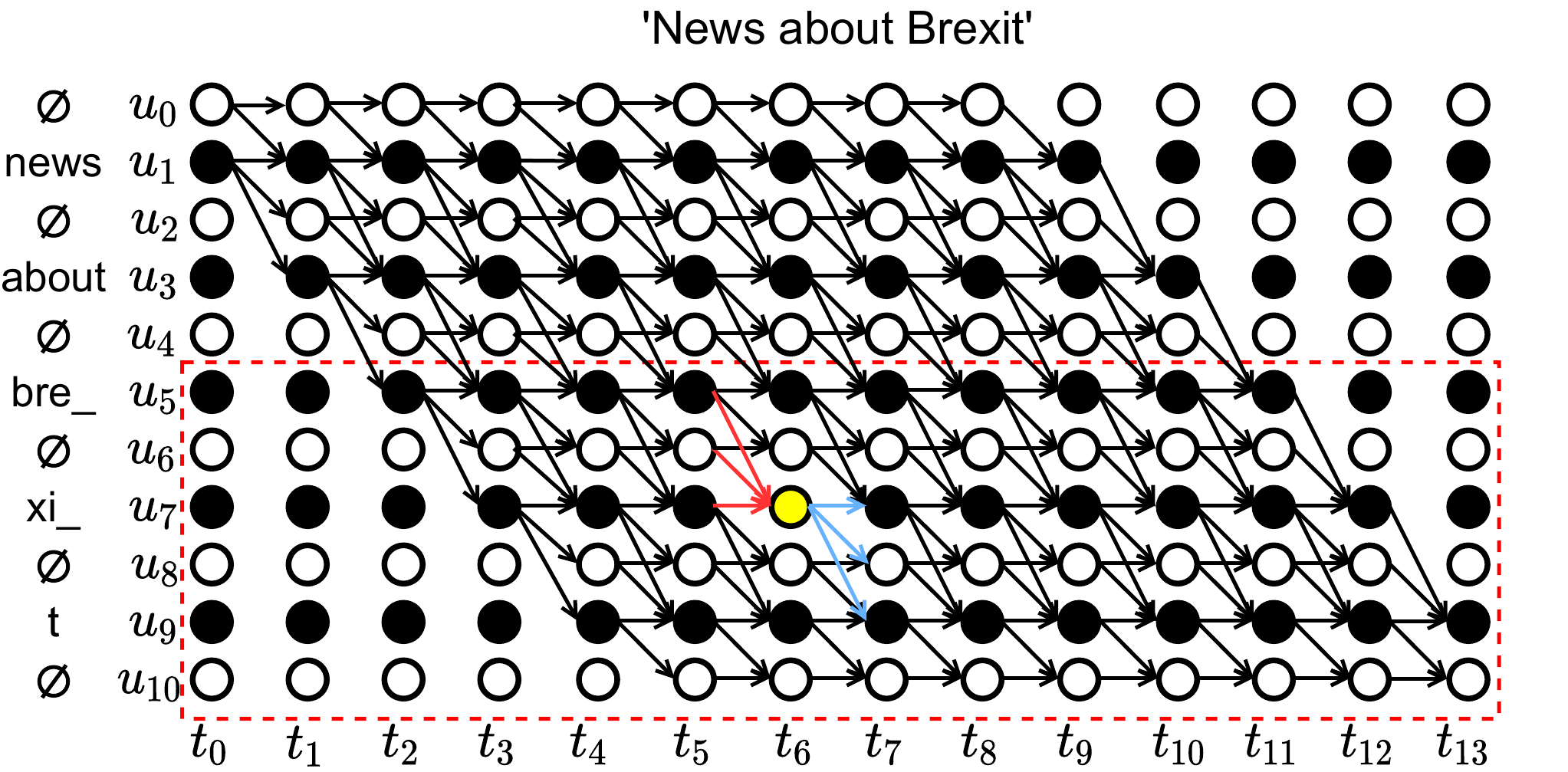}  
\caption{Illustration of CTC decoding lattices for the example sentence of 'News about Brexit', where the modeling unit is subword and ``Brexit" is an OOV word. Black nodes are label tokens and white nodes are blank tokens.}
\label{fig:ctc-decoding}
\end{figure}

\begin{figure}[!t]
\centering
\includegraphics[width=5cm]{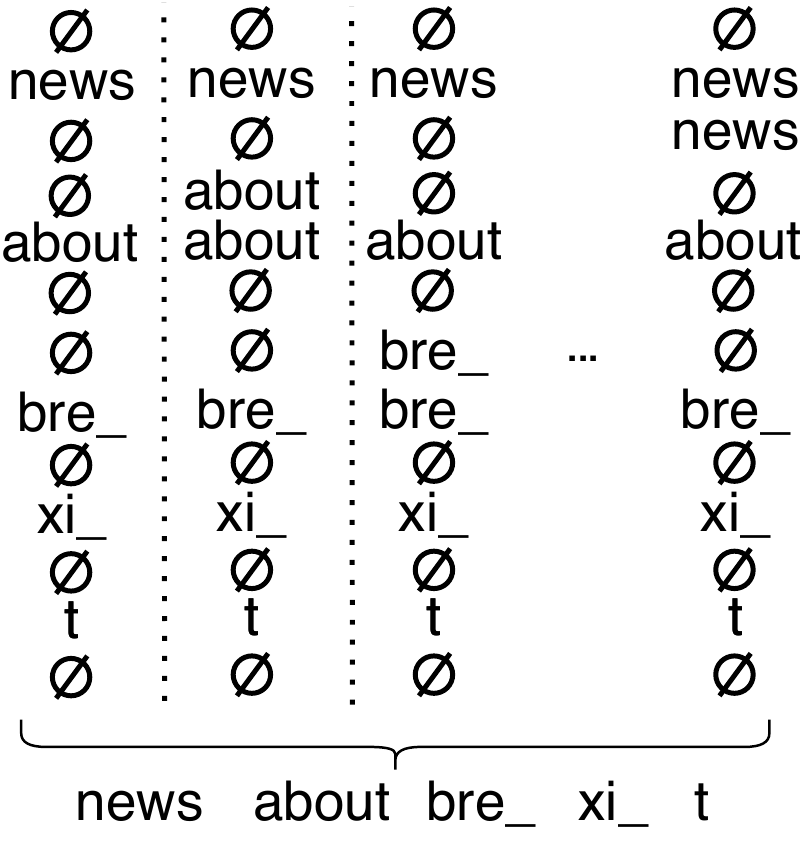}  
\caption{Possible intermediate decoding sequences of the CTC function.}
\label{fig:ctc-intermediate}
\end{figure}

After processing the last input vector $x_{T}$, we can get a $(2U+1) \times T$ lattice matrix, for instance the one in Figure~\ref{fig:ctc-decoding} for the utterance ``News about Brexit" modeling on subword units is shown. Figure~\ref{fig:ctc-intermediate} lists some intermediate decoding results of the CTC function. The final CTC loss contains the sum of probabilities of all possible paths that can be converted to be the target labels $\textbf{y}$ by merging repeat units and removing blank tokens as shown in Eq.~\eqref{word1}, where $\textbf{a}$ is a possible token path and $\mathcal{F}:\widetilde{\textbf{y}}\rightarrow\textbf{y}$ is the function that maps the extended label sequence $\widetilde{\textbf{y}}$ back to the true label sequence $\textbf{y}$. 

\begin{figure}[!t]
\centering
\includegraphics[width=9cm]{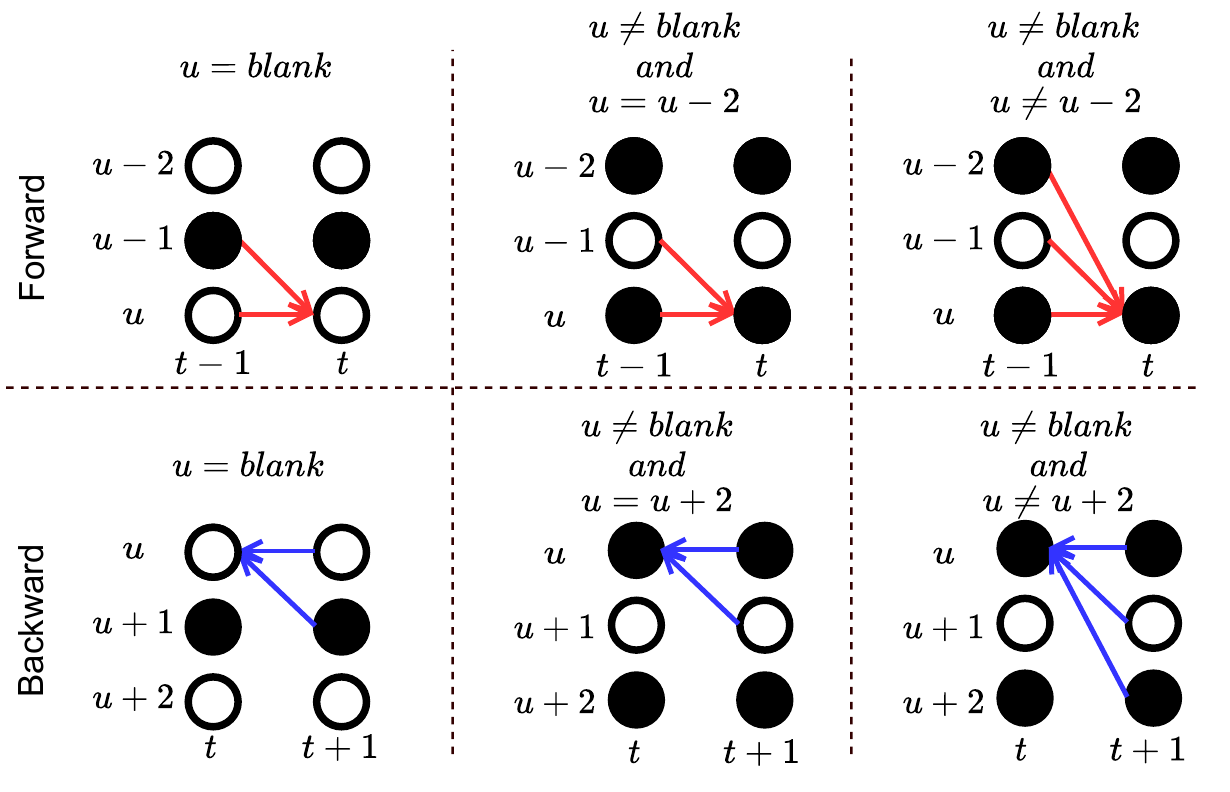}  
\caption{Diagram of three cases in forward or backward computation, where black nodes are label tokens and white nodes are blank tokens. Red and blue arrows are forward and backward paths respectively.}
\label{fig:forward-backward}
\end{figure}

 We denote $\widehat{y}(t,u)$ and $b(t,u)$ as the probability of a label and a blank token at node $(t,u)$ respectively. According to the definition of CTC~\cite{kawakami2008supervised}, as shown in Figure~\ref{fig:forward-backward}, for any blank tokens, there are only two paths from the previous step reaching the current blank token since label tokens can not be skipped. When the current node is a label token and is same as $u-2$, there is no connection from $u-2$ to $u$. A blank token has to be generated to separate the adjacent same tokens. For instance, there is no path from the first `o' to the second one when decoding the word `look'. Lastly, three possible paths can reach the label token when $u \neq u-2$. Three same cases for the backward procedure are shown in Figure~\ref{fig:forward-backward} as well.
 

 According to the three cases shown in Figure~\ref{fig:forward-backward}, the forward variable $\alpha(t,u)$ can be calculated recursively as follows:

\fontsize{8.0pt}{10.5pt}\selectfont\begin{equation}
\label{word2}
\alpha(t,u) =\left\{\begin{array}{ll}\widehat{y}(t-1,u-1)\alpha(t-1,u-1) &{\text{if } (t,u) = blank}\\
+b(t-1,u)\alpha(t-1,u)&\\
b(t-1,u-1)\alpha(t-1,u-1) & \text{if } (t,u) \neq blank \\ 
+\widehat{y}(t-1,u)\alpha(t-1,u)&\text{ and (t,u)} = ({\rm t}-1, {\rm u}-2)\\
\widehat{y}(t-1,u-2)\alpha(t-1,u-2) &\text{if } (t,u)\neq blank \\ 
+b(t-1,u-1)\alpha(t-1,u-1)&\text{ and } (t,u) \neq (t-1, u-2)\\
+\widehat{y}(t-1,u)\alpha(t-1,u)&
\end{array}\right.
\end{equation}\normalsize







It is worth noting that when the two adjacent tokens are same, i.e. $(t,u) = (t-1,u-2)$, there is no direct transition between the two repeated tokens and the second one can only be reached through the blank token or the node of $(t-1,u)$.

Similarly, blue arrows reaching the node $(t,u)$ and the backward variable $\beta(t,u)$ can be represented by Eq.~\eqref{word3} in three cases.

\fontsize{8pt}{10.5pt}\selectfont\begin{equation}
\label{word3}
\beta(t,u) =\left\{\begin{array}{ll}b(t,u)\beta(t+1,u) &\text{if } (t,u) = blank\\
+\widehat{y}(t,u)\beta(t+1,u+1) &\\
\widehat{y}(t,u)\beta(t+1,u) & \text{if } (t,u) \neq blank \\ 
+b(t,u)\beta(t+1,u+1)&\text{ and } (t,u) = (t+1, u+2)\\
\widehat{y}(t,u)\beta(t+1,u) &\text{if } (t,u)\neq blank \\ 
+b(t,u)\beta(t+1,u+1)&\text{ and } (t,u) \neq (t+1, u+2)\\
+\widehat{y}(t,u)\beta(t+1,u+2)&
\end{array}\right.
\end{equation}\normalsize




$P(\mathcal{A}_{t,u}|\textbf{x})$, the probability of any candidate paths passing through node $(u,t)$ conditioned on the input sequence $\textbf{x}$, can be obtained by multiplying forward (Eq.~\eqref{word2}) and backward probabilities (Eq.~\eqref{word3}) by Eq.~\eqref{word4}. 


\begin{eqnarray}
\label{word4}
P(\mathcal{A}_{t,u}|\textbf{x})&=&\alpha(t,u)\beta(t,u)
\end{eqnarray}

Thereby, the gradient of the CTC loss function $\mathcal{L}_{CTC}$ \textit{$w.r.t$} $\widehat{y}(t,u)$ and $b(t,u)$ can be estimated by Eq.~\eqref{word5} and Eq.~\eqref{word6} respectively.





\fontsize{7.7pt}{10.5pt}\selectfont\begin{equation}
\label{word5}
\frac{\partial \mathcal{L}_{CTC}}{\partial \widehat{y}(t,u)}\propto \left\{\begin{array}{ll}
\alpha(t,u)\beta(t+1,u+1)& \text{if } (t,u) = blank\\
\alpha(t,u)\beta(t+1,u)&\text{if } (t,u) \neq blank\\
&\text{ and } (t,u) = (t+1, u+2)\\
\alpha(t,u) \text{(}\beta(t+1,u) &\text{if } (t,u)\neq blank \\
+\beta(t+1,u+2)\text{)}&\text{ and } (t,u) \neq (t+1, u+2)\\
\end{array}\right.
\end{equation}\normalsize

\begin{equation}
\label{word6}
\frac{\partial \mathcal{L}_{CTC}}{\partial b(t,u)}\propto \left\{
\begin{array}{ll}
\alpha(t,u)\beta(t+1,u) &\text{if } (t,u) = blank\\
\specialrule{0em}{0.5ex}{1.0ex}

\alpha(t,u)\beta(t+1,u+1) &\text{if } (t,u) \neq blank\\

\end{array}\right.
\end{equation}

The CTC function treats all nodes equally and aims to minimize the global loss, which makes models hardly focus on local connections in the decoding lattice. To guide models to pay more attention to the OOV words, we emphasize the OOV words (the nodes in the dotted box in Figure~\ref{fig:ctc-decoding}) by rescaling the probabilities of OOV nodes in candidate alignments. Thus, the rescaled probability of all alignments passing through OOV nodes is as follows:

\begin{equation}
\label{word7}
\widetilde{P}(\mathcal{A}_{t,u}|\textbf{x}) =
\begin{cases}
\mu P(\mathcal{A}_{t,u}|\textbf{x}), &\text{if } u \in O\\
P(\mathcal{A}_{t,u}|\textbf{x}), &\text{otherwise}
\end{cases}
\end{equation}

The regularized loss function at word level is:

\begin{equation}
\label{word8}
\mathcal{L}_{word}=-\sum_{\mathcal{A}\subset\mathcal{F}^{-1}(\widetilde{\textbf{y}})}\widetilde{P}(\mathcal{A}_{t,u}|\textbf{x})
\end{equation}

We implement our approach by multiplying the gradients of OOV nodes on the candidate path with $\mu$, as shown in the following equations:

\begin{equation}
\label{word9}
\frac{\partial \mathcal{L}_{word}}{\partial \widehat{y}(t,u)} =
\begin{cases}
\mu \frac{\partial \mathcal{L}_{CTC}}{\partial \widehat{y}(t,u)}, &\text{if } u \in O\\
\frac{\partial \mathcal{L}_{CTC}}{\partial \widehat{y}(t,u)}, &\text{otherwise}
\end{cases}\\
\end{equation}

\begin{equation}
\frac{\partial \mathcal{L}_{word}}{\partial b(t,u)} =
\begin{cases}
\mu \frac{\partial \mathcal{L}_{CTC}}{\partial b(t,u)}, &\text{if } u \in O\\
\frac{\partial \mathcal{L}_{CTC}}{\partial b(t,u)}, &\text{otherwise}
\end{cases}\\
\end{equation}
where $O$ is the set of OOV words tokenized into subwords.

 \subsection{Overcoming Catastrophic Forgetting}

 Directly fine-tuning models on a dataset obeying a different distribution from the original training set may lead to catastrophic forgetting. The updated model may overfit the new dataset but forget the knowledge learned on the original one. To overcome models suffering from catastrophic forgetting, we adopt two approaches during fine-tuning. The first one is mixing partial original audio from LibriSpeech used for baseline model training with synthetic speech, since adding data that obeys the same distribution as the training set can efficiently mitigate the forgetting problem. We explore the effect of different mixing ratios and present the results in Section~\ref{experimental results}. The other approach is constraining model parameters from updating during fine-tuning with L2 regularization or EWC~\cite{kirkpatrick2017overcoming}, and we will introduce the details in the following sections.
 
 \subsubsection{L2 Regularization}
  The L2 regularization loss $\mathcal{L}_{L2}(\theta)$ is shown in Eq.~\eqref{L2}, where $\mathcal{L}(\theta)$ is the original CTC loss or rescaled loss in Eq.~\eqref{sentence level} and Eq.~\eqref{word8}. $\theta_{i}$ is the $i$th parameter of the ASR model to be updated during fine-tuning, and $\theta^{'}_{i}$ is the $i$th parameter in the baseline model which is invariable and saved locally. $\lambda$ is the coefficient to balance the scale of two parts. 
 
\begin{equation}
\mathcal{L}_{L2}(\theta)=\mathcal{L}_{CTC}(\theta)+\frac{\lambda}{2}\sum_{i}^{}(\theta_{i}-\theta^{'}_{i})^{2}
\label{L2}
\end{equation}

L2 loss takes the difference between the fine-tuned model and the old model into account to ensure the updated model will not stray away too much from the baseline.

\subsubsection{Elastic Weight Consolidation}
 
 Different from L2 loss that always refers to a fixed standard and treats all parameters equally, the EWC loss as shown in Eq.~\eqref{EWC} uses the diagonal of the Fisher information matrix $F$ to dynamically weigh the importance of each model parameter for the source task.
 
\begin{equation}
\mathcal{L}_{EWC}(\theta)=\mathcal{L}_{CTC}(\theta)+\frac{\lambda}{2}\sum_{i}^{}F_{i}\cdot(\theta_{i}-\theta^{'}_{i})^{2}
\label{EWC}
\end{equation}

The Fisher information matrix $F$ can be estimated by the following equation with the gradients of the convergent source model.


\begin{equation}
F_{i}=\frac{1}{|D|}\sum_{d \in D}^{}\frac{\partial^2  \mathcal{L}_{CTC}(d,{\theta^{'}_{i}}) }{\partial {\theta^{'}_{i}}^2}\approx \frac{1}{|D|}\sum_{d \in D}^{}\frac{\partial  \mathcal{L}_{CTC}(d,{\theta^{'}_{i}})^2 }{\partial {\theta^{'}_{i}}^{2}}
\label{f-matrix}
\end{equation}

\noindent where $\theta^{'}$ and $D$ are the parameters and the dataset used in the source ASR task respectively. The more important parameters for the source task would have the larger values in the Fisher information matrix, which constrains the change of important parameters and avoids knowledge forgetting. The less important ones would have relatively small values and are encouraged to adapt on the new task. In this paper, the diagonal of the Fisher information matrix is estimated on the LibriSpeech 960h training set.

 \section{Experiments}
 
 \subsection{ASR Model Architecture}
 \label{sec:u2}
\begin{figure}[!t]
  \centering
  \includegraphics[width=8cm]{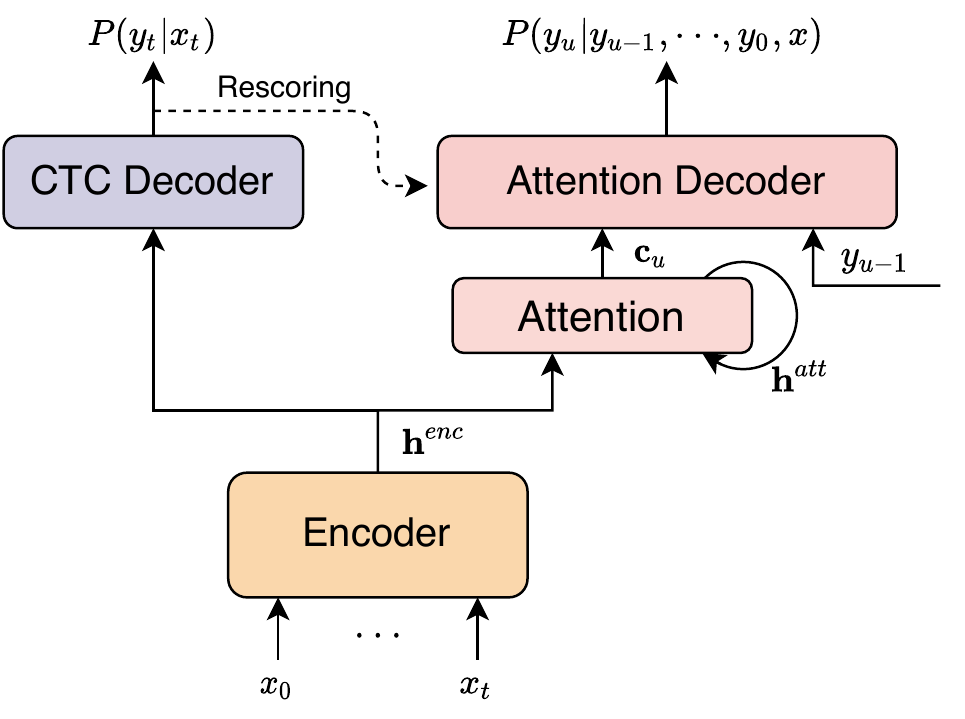}  
  \caption{Two-pass hybrid CTC/attention ASR architecture.}
  \label{fig:hybird ctc-attention}
\end{figure}

The end-to-end ASR model used in our experiments is the two-pass hybrid CTC/attention architecture, U2~\cite{zhang2020unified}, as shown in Figure ~\ref{fig:hybird ctc-attention}. The shared encoder converts acoustic features $x$ into a latent vector $\textbf{h}^{enc}$, then the CTC decoder transforms the latent vector into character/word probability $P(y_{t}|x_{t})$ with the same length as the input frames. Meanwhile, the attention decoder generates one character/word probability $P(y_{u}|y_{u-1}, \cdot \cdot \cdot,y_{0}, x)$ per time step by conditioning on the attention content vector $\textbf{c}_{u}$ and the decoder output from the last step $y_{u-1}$. During training, the sum of CTC loss and attention loss is used to do back-propagation, while during inference, the n-best hypotheses produced by the CTC decoder are rescored by the attention decoder to obtain better performance. The candidate with the highest score will be the final output. 

We develop our model based on the U2 model published with the WeNet toolkit~\cite{zhang2021wenet} which unifies streaming and non-streaming ASR models into one architecture by proposing the dynamic chunk-based attention. The encoder consists of 12 conformer blocks, with 4 multi-head attention, 2048 linear units, swish activation, a positional dropout rate of 0.1 and Conv2D kernel size of 31 for each block. The attention decoder contains 6 transformer blocks, and the CTC decoder is composed of 1 linear layer and 1 log softmax function. The U2 model is pre-trained on the 960h LibriSpeech corpus and achieves 3.18\% of WER and 8.72\% of WER on the test-clean and the test-other test set respectively when rescoring with attention decoder. The pre-trained weights are available on the website\footnote{http://mobvoi-speech-public.ufile.ucloud.cn/public/wenet/librispeech/20210216\_\\conformer\_exp.tar.gz}, which will be the baseline model for all our experiments.
 
 \subsection{OOV Set with Real Audio}
 We build a 100-OOV-word dataset from LRS3-TED~\cite{afouras2018lrs3} corpus since there are no standard OOV corpora published by the community. LRS3-TED is an audio-visual dataset collected from TED and TEDx talks with spontaneous speech and various speaking styles. It is comprised of over 400 hours of video by more than 5000 speakers and contains an extensive vocabulary. We filter the vocabulary existing in LRS3-TED but not present in LibriSpeech and select 100 OOV words from more than 100 speakers used for test, where each OOV word contains 50 utterances. We random split these utterances into training, validation and test sets with a ratio of 2:1:2. The duration ratio of training, validation and test sets is 3h : 1.6h : 2.8h. In the rest of this paper, the three sets will be referred to as real OOV (ROOV) training, ROOV val and ROOV test set respectively. We report all experimental results on the ROOV test set. More details about the 100 OOV words can be found in the Appendix.

  \subsection{OOV Set with Synthetic Audio}
  
  Our goal in this paper is to improve OOV word recognition with synthetic audio and loss rescaling methods. In this subsection, we introduce the synthetic dataset used for model training. 
  
 \textbf{Text Crawling}: we crawl 100 sentences for each new word with Scrapy\footnote{https://scrapy.org/}, where each sentence contains less than 50 words in case of running out of memory. During crawling, we filter those sentences include OOV words out of the selected 100-OOV-word set to ensure that the model performance is only influenced by the chosen ones.
 
 \textbf{Speech Synthesis}: we split the 100 sentences for each new word into training and validation with a ratio of 9:1. The model evaluation will be conducted on the ROOV test set with real audio. Instead of applying a single multi-speaker TTS system, we use several commercial Application Programming Interfaces (APIs), i.e.\ Baidu TTS API\footnote{https://ai.baidu.com/tech/speech/tts}, Google TTS API\footnote{https://cloud.google.com/text-to-speech}, iFLYTEK TTS API\footnote{https://www.xfyun.cn/doc/tts/online\_tts/API.html}, Tencent TTS API\footnote{https://cloud.tencent.com/document/api} and Alibaba TTS API\footnote{https://www.alibabacloud.com/help/doc-detail/84435.htm} to synthesize audio and to enable more voices and more variety in speech. In contrast to open-source multi-speaker TTS models, the commercial APIs produce higher quality speech, which is crucial for the following experiments. 8 different speaker voices (4 males and 4 females) are used for training set synthesis and 2 voices (1 male and 2 female) for the validation set. There is no voice overlap between the 2 parts. The duration ratio of synthetic audio used for training and validation is 9.0h : 2.3h. In the rest of this paper, the synthetic data will be referred to as synthetic OOV (SOOV) training and SOOV val set respectively.
 
 \textbf{Data Augmentation}: to avoid overfitting, we perturb speech on speed with the factors of 0.9, 1.0, and 1.1. Furthermore, clean speech is augmented with 5 kinds of room impulse responses\footnote{http://www.iks.rwth-aachen.de/en/research/tools-downloads/databases/aachen-impulse-response-database/}. 10 noise sources~\cite{reddy2019scalable}, such as announcements, appliances, and traffic, are added to the reverberated speech with 6 levels of speech-to-noise ratio (0, 4, 8, 12, 16 and 20). Moreover, SpecAugment~\cite{park2019specaugment} with 2 frequency masks (maximum width 50) is utilized on the fly during training. 

\subsection{Evaluation Metrics}
We use 3 metrics to evaluate the experimental results of our proposed method:
\begin{itemize}
\item \textbf{WER}: word error rate is the ratio of error terms, i.e., substitutions, deletions, and insertions, to the total number of words in the reference.
\item \textbf{Recall}: recall is the number of true positives $TP$ over the sum of the number of true positives and the number of false negatives $FN$.
\begin{equation}
\label{recall}
Recall=\frac{TP}{TP+FN}
\end{equation}

\item \textbf{Precision}: precision is the number of true positives $TP$ over the sum of the number of true positives and the number of false positives $FP$.
\begin{equation}
\label{recall}
Precision=\frac{TP}{TP+FP}
\end{equation}

\end{itemize}

 \subsection{Training Settings}
 The baseline model is trained on the 960h LibriSpeech dataset with a batch size of 12, an initial learning rate of 4e-3 and 25000 warm-up steps. When doing fine-tuning for OOV experiments, we use a bigger batch size of 20 to enable the model to see more utterances not containing OOV words and avoid loss explosion. A tiny initial learning rate of 4e-6 is utilized for fine-tuning, which is annealed with a value of 1.1 after every 3000 steps, since a tiny learning rate can efficiently ensure stable model learning and retain the previously learned knowledge. In addition, to avoid gradient explosion, we clip all gradients greater than 2, while the threshold used in baseline training is 5.
 
 The validation set during model training is the mixture of the LibriSpeech dev and OOV TTS dev set with the ratio of 1:1. The model checkpoint performing best on the mixture validation set is used for evaluation on test sets with early stopping. It is noteworthy that the attention mechanism and attention decoder are always frozen since we found that fine-tuning the entire network leads to gradient explosion. In addition, it is hard to balance the CTC loss and attention loss since the CTC loss is rescaled in our methods. We only use the the attention and the decoder for rescoring.

 \section{Experimental Results}
 \label{experimental results}
 In this section, we report the experimental results from the following perspectives.
 
\subsection{Results of Speech Mixture from Source and Target Domain}
 
 To mitigate catastrophic forgetting, we mix original real speech in the LibriSpeech training set (source domain) with the ROOV or the SOOV training set (target domain). It is still an open question what the best mixing ratio is. We fine-tune the baseline model with different ratios and report results on the standard LibriSpeech test sets (test-clean and test-other) and the ROOV test set. 

 As shown in Table~\ref{tab:ratio-lr}, when fine-tuning only with ROOV training set (real speech from LRS3-TED), the model shows the inability to retain old knowledge and performs badly on previous LibriSpeech tasks, which leads to a tremendous rise in WER on the test-other set from 8.72\% to 41.52\%. Context information is crucial for ASR models. In the 0:1 setting, the pre-trained ASR model is destroyed, especially the learned context knowledge. Consequently, the overfitted model is hard to infer a correct context and recognize OOV words.

  \begin{table}[th]
  \setlength{\abovecaptionskip}{0.5cm}
  \setlength{\belowcaptionskip}{0.4cm}
  \caption{The influence of the ratio of speech data from LibriSpeech and real speech in ROOV training set (LRS3-TED) on ASR and OOV word recognition.}
  \label{tab:ratio-lr}
  \centering
  \resizebox{12cm}{!}{
  \normalsize 
\begin{tabular}{ccccccc}
\toprule[1.5pt]
\multirow{2}{*}{\textbf{Model}} & \multirow{2}{*}{\textbf{Ratio}} &\textbf{WER} &\textbf{WER} &\textbf{WER} & \textbf{Recall} & \textbf{Precision} \\ \cline{3-7} 
 &  & \textbf{test-clean} & \textbf{test-other} & \textbf{ROOV test}& \textbf{ROOV test}& \textbf{ROOV test} \\ \midrule[1.5pt]
\textbf{Baseline} & - & 3.18 & 8.72 & 15.33 & 1.37 & 100 \\ \hline
\multirow{5}{*}{\textbf{{\begin{tabular}[c]{@{}c@{}}LibriSpeech\\+\\ROOV training\end{tabular}}}} & 0:1 & 30.18 & 41.52 & 38.21 & 19.58 & 97.28 \\ \cline{2-7} 
 & 1:1 & 23.34 & 30.41 & 25.83 & 24.30 & 93.18 \\ \cline{2-7} 
 & 2:1 & 13.26 & 26.77 & 23.27 & 32.05 & 98.41 \\ \cline{2-7} 
 & 3:1 & 7.26 & 15.25 & 22.31 & 30.82 & 98.02 \\ \cline{2-7} 
 & 4:1 & 4.11 & 11.05 & 15.52 & 26.12 & 99.04 \\
 \bottomrule[1.5pt]
\end{tabular}}
\end{table}

 The forgetting tendency slows down as original data is incorporated into training. When the ratio of audio from LibriSpeech and LRS3-TED is 2:1, the model achieves the highest recall of 32.05\%. The more data from the source task (LibriSpeech) is used for training, the more previous knowledge is retained and the better performance is obtained on the LibriSpeech evaluation sets. However, the model tends to focus on the previous LibriSpeech tasks as the ratio increases, which leads to the decrease of recall on the ROOV test. 

We can draw the same conclusion when fine-tuning ASR models with synthetic data as shown in Table~\ref{tab:ratio-ls}. We prioritize the model performance regarding recall since the goal of this paper is to enable the ASR model to learn new vocabulary, and the catastrophic forgetting issue will be tackled in the next section. Therefore, the 2:1 mixture ratio is used in the following experiments.

  \begin{table}[th]
  \setlength{\abovecaptionskip}{0.5cm}
  \setlength{\belowcaptionskip}{0.4cm}
  \caption{The influence of the ratio of speech data from LibriSpeech and synthetic speech in SOOV training set on ASR and OOV word recognition.}
  \label{tab:ratio-ls}
  \centering
  \resizebox{12cm}{!}{
  \normalsize 
\begin{tabular}{ccccccc}
\toprule[1.5pt]
\multirow{2}{*}{\textbf{Model}} & \multirow{2}{*}{\textbf{Ratio}} &\textbf{WER} &\textbf{WER} &\textbf{WER} & \textbf{Recall} & \textbf{Precision} \\ \cline{3-7} 
 &  & \textbf{test-clean} & \textbf{test-other} & \textbf{ROOV test}& \textbf{ROOV test}& \textbf{ROOV test} \\ \midrule[1.5pt]
\textbf{Baseline} & - & 3.18 & 8.72 & 15.33 & 1.37 & 100 \\ \hline
\multirow{5}{*}{\textbf{{\begin{tabular}[c]{@{}c@{}}LibriSpeech\\+\\SOOV training\end{tabular}}}} & 0:1 & 35.54 & 53.42 & 42.27 & 13.51 & 99.83 \\ \cline{2-7} 
 & 1:1 & 29.05 & 39.08 & 30.22 & 20.51 & 91.21 \\ \cline{2-7} 
 & 2:1 & 20.34 & 28.72 & 25.31 & 27.54 & 97.67 \\ \cline{2-7} 
 & 3:1 & 13.37 & 19.21 & 23.22 & 26.25 & 98.53 \\ \cline{2-7} 
 & 4:1 & 6.71 & 16.23 & 20.31 & 23.21 & 98.08 \\
 \bottomrule[1.5pt]
\end{tabular}}
\end{table}

\subsection{Results of Loss Rescaling at Sentence Level}

In this section, we explore the effect of loss rescaling at sentence level. We compare the model performance using real speech data with the results using synthetic audio. As shown in Table~\ref{tab:sentence-level}, using L2 and EWC regularization efficiently reduces catastrophic forgetting and improves the recall rate on OOV words while the WER increases only in few cases on the test-clean and test-other test sets. We find $\lambda=5e7$ is the best weight to balance the L2/EWC loss and the ASR losses. 

We reproduce the method proposed by Zheng et al.~\cite{zheng2021using}, in which a RNN-T ASR model is fine-tuned with EWC on mixed real and synthetic audio. However, the dataset is not published. It is noteworthy that the experimental results reported in Table~\ref{tab:sentence-level} and Table~\ref{tab:word-level} are based on our generated data with the method proposed by Zheng et al.~\cite{zheng2021using}. In addition, as shown in Table~\ref{tab:sentence-level}, row ``Isolated Words", we de-emphasize the non-OOV words by fine-tuning ASR models with utterances containing only isolated OOV words which are segmented from real or synthetic continuous speech according to the time alignment information obtained from the Montreal Forced Aligner\footnote{https://github.com/MontrealCorpusTools/Montreal-Forced-Aligner}. Compared with the method proposed by Zheng et al.~\cite{zheng2021using}, fine-tuning ASR models with only isolated OOV words can effectively improve the recall rate but it leads to much more serious forgetting on non-OOV recognition.

 \begin{table}[th]
  \setlength{\abovecaptionskip}{0.5cm}
  \setlength{\belowcaptionskip}{0.4cm}
  \caption{Loss rescaling at sentence level with L2/EWC regularization. $\mu$ and $\lambda$ are the loss weight in Eq.~\eqref{sentence level} and the weight of L2/EWC in Eq.~\eqref{L2}/Eq.~\eqref{EWC} respectively. R is short for ``Real" and represents using ROOV training set with real audio for fine-tuning. S is short for ``Synthetic" and denotes using SOOV training set with synthetic audio for fine-tuning.}
  
  \label{tab:sentence-level}
  \centering
  \resizebox{12cm}{!}{
\begin{tabular}{ccc|cc|cc|cc|cc|cc}
\toprule[1.5pt]
 &
  \textbf{$\mu$} &
  \textbf{$\lambda$} &
  \multicolumn{2}{c}{\textbf{WER$\downarrow$}} &
  \multicolumn{2}{c}{\textbf{WER$\downarrow$}} &
  \multicolumn{2}{c}{\textbf{WER$\downarrow$}} &
  \multicolumn{2}{c}{\textbf{Recall$\uparrow$}} &
  \multicolumn{2}{c}{\textbf{Precision$\uparrow$}} \\ \hline
\textbf{Test sets} &
  - &
  - &
  \multicolumn{2}{c|}{\textbf{test-clean}} &
  \multicolumn{2}{c|}{\textbf{test-other}} &
  \multicolumn{2}{c|}{\textbf{ROOV test}} &
  \multicolumn{2}{c|}{\textbf{ROOV test}} &
  \multicolumn{2}{c}{\textbf{ROOV test}} \\ \hline
  
\textbf{Model}                & -     & -   & R   & S & R & S & R     & S & R     & S & R     & S \\ \toprule[1.5pt]
\textbf{Baseline} &
  1 &
  0 &
  \multicolumn{2}{c|}{3.18} &
  \multicolumn{2}{c|}{8.72} &
  \multicolumn{2}{c|}{15.33} &
  \multicolumn{2}{c|}{1.37} &
  \multicolumn{2}{c}{100} \\ \hline
\multirow{5}{*}{\textbf{L2}}  & 1     & 5e7 & 5.21 & 5.32  & 10.22  & 10.91  &  15.03    & 18.84  & 24.13 & 15.24  & 98.12 &  99.02 \\ \cline{2-13} 
                              & 10    & 5e7 &  5.53    & 5.36  & 10.93&  12.02 & 15.17 &  18.69 & 35.12 &  24.42 & 98.51 &  92.03 \\ \cline{2-13} 
                              & 100   & 5e7 &  6.07    &  6.38 &    11.38  &  11.94 &  15.01     &  19.28 & 50.02 &  31.26 & 94.35 & 95.71  \\ \cline{2-13} 
                              & 1000  & 5e7 &  6.83    &  6.93 &   11.83   & 12.48  &  15.74     &  19.74 & 53.42 &  39.72 & 90.04 & 89.43  \\ \cline{2-13} 
                              & 10000 & 5e7 &  7.22    &  7.79 &   13.48   &  13.59 &   16.33    &  19.26 & 51.78 &  39.77 & 83.44 & 78.25  \\ \toprule[1.5pt]
\textbf{Zheng et al.~\cite{zheng2021using}} &
  1 &
  5e7 &
 3.18  &
 3.20  &
  8.93 &
  9.02 &
  14.87 &
  18.23 &
  30.57 &
  16.20 &
  98.43 & 99.12 \\ \toprule[0.5pt]
\textbf{Isolated Words} & 1 & 5e7 &   5.83   &  6.32 &  11.01    &  12.27 &   14.65    & 18.47  & 35.92 & 22.48  & 97.65 & 98.15  \\ \toprule[0.5pt]

\multirow{4}{*}{\textbf{EWC}} & 10    & 5e7 &   5.21   &  5.20 &  9.91    &  9.83 &   14.59    & 18.34  & 49.22 & 30.71  & 98.11 & 99.04  \\ \cline{2-13} 
                              & 100   & 5e7 &    5.29  &  5.37 &    9.82  &  9.98 &  14.66      & 18.04  & 54.28 &  41.22 & 97.21 & 97.55  \\ \cline{2-13} 
                              & 1000  & 5e7 &    5.55  &  5.54 &    10.54  &  11.37 &   14.67    & 18.37  & 54.06 &  42.38 & 88.24 & 89.41  \\ \cline{2-13} 
                              & 10000 & 5e7 &   6.07   &  6.12 &    11.78  &  11.84 &   14.52    & 18.02  & 55.31 &  42.58 & 80.51 & 84.28\\
\bottomrule[1.5pt]
\end{tabular}} 
\end{table}

When just fine-tuning the base model with real or synthetic OOV audio, all words are treated equally, which leads the model to hardly focus on the OOV words we concern. Therefore, we propose loss rescaling and encourage the model to pay more attention to OOV words by enlarging the loss of sentences containing unknown words. For the loss rescaling weight $\mu$, we examine the values 1, 10, 100, 1000, and 10,000. As we can see in Table~\ref{tab:sentence-level}, the OOV recall rapidly increases when rescaling the target sentences by 100 times compared to only fine-tuning using L2 (50.02\% VS 24.13\% for real speech and 31.26\% VS 24.13\% for synthetic audio). 

As a bigger $\mu$ is used, the recall further rises, but the WER on the test-clean and the test-other test sets is getting worse. We hypothesize that directly rescaling the entire sentence loss may also enhance irrelevant words or noises, which leads to gradient explosion during training and accelerates forgetting previous knowledge. Hence, we have to use a very small learning rate and clip the gradients over 2.0 to ensure the progress of fine-tuning. In contrast to L2 regularization, EWC can provide more stable and resilient protection of the weights important for the previous LibriSpeech tasks but still with a relatively high loss in the ASR performance, as shown in Table~\ref{tab:sentence-level}. 

In addition, when fine-tuning only with the synthetic audio, we obtain competitive recall rate compared to utilizing real speech from the LRS3-TED dataset, for example, when using EWC and rescaling the loss by 1000 times larger, we achieve 54.06\% VS 42.38\% recall rate for real and synthetic speech. Furthermore, compared with the method proposed by Zheng et al.~\cite{zheng2021using}, rescaling the OOV utterance loss can achieve significant improvement on recall with only slight decrease on WER and precision. 
 
\subsection{Results of Loss Rescaling at Word Level}

Instead of enhancing the entire sentence loss, in this section, we report the results of only rescaling unknown words. As shown in Table~\ref{tab:word-level}, the $\lambda$ weight, which is needed to balance the ASR and L2/EWC loss, is smaller (1e7) than the one used at sentence level (5e7) in Table~\ref{tab:sentence-level}, and we do not observe gradient explosion during training unless $\mu$ is very large, e.g.\ 1e4. The results without loss rescaling ($\mu = 1$) is slightly different from the one shown in Table~\ref{tab:sentence-level}, which is caused by different $\lambda$ values used for L2/EWC normalization.
Using 10 times smaller $\mu$ can obtain a similar or even higher recall at word-level rescaling, for example, using $\mu$ of 100 gets a 45.81\% recall rate at word level compared to using $\mu$ of 1000 getting a 42.38\% recall rate at sentence level when regularizing models with EWC and fine-tuning with only synthetic data.

Rescaling loss on OOV words obtains lower WER on the standard LibriSpeech test sets. Moreover, the bigger $\mu$ is used, the higher recall is achieved for the ROOV test. Furthermore, a worse WER is obtained for the LibriSpeech benchmark, which is observed at sentence-level loss rescaling as well. To make a trade-off between WER and Recall, 100 times loss rescaling is performing best for our experimental results with only a 7.8\%/4.6\% relative WER increase on the test-clean/test-other test set or a 45.81\% recall rate on the ROOV test. In addition, we obtain competitive performance
by only using synthetic audio compared with using real speech data for fine-tuning.

\begin{table}[th]
  \setlength{\abovecaptionskip}{0.5cm}
  \setlength{\belowcaptionskip}{0.4cm}
  \caption{Loss rescaling at word level with L2/EWC regularization. $\mu$ and $\lambda$ are the loss weight in Eq.~\eqref{sentence level} and the weight of L2/EWC in Eq.~\eqref{L2}/Eq.~\eqref{EWC} respectively. R is short for ``Real" and represents using ROOV training set with real audio for fine-tuning. S is short for ``Synthetic" and denotes using SOOV training set with synthetic audio for fine-tuning.}
  \label{tab:word-level}
  \centering
  \resizebox{12cm}{!}{
\begin{tabular}{ccc|cc|cc|cc|cc|cc}
\toprule[1.5pt]
 &
  \textbf{$\mu$} &
  \textbf{$\lambda$} &
  \multicolumn{2}{c}{\textbf{WER$\downarrow$}} &
  \multicolumn{2}{c}{\textbf{WER$\downarrow$}} &
  \multicolumn{2}{c}{\textbf{WER$\downarrow$}} &
  \multicolumn{2}{c}{\textbf{Recall$\uparrow$}} &
  \multicolumn{2}{c}{\textbf{Precision$\uparrow$}} \\ \hline
\textbf{Test sets} &
  - &
  - &
  \multicolumn{2}{c|}{\textbf{test-clean}} &
  \multicolumn{2}{c|}{\textbf{test-other}} &
  \multicolumn{2}{c|}{\textbf{ROOV test}} &
  \multicolumn{2}{c|}{\textbf{ROOV test}} &
  \multicolumn{2}{c}{\textbf{ROOV test}} \\ \hline
  
\textbf{Model}                & -     & -   & R   & S & R & S & R     & S & R     & S & R     & S \\ \toprule[1.5pt]
\textbf{Baseline} &
  1 &
  0 &
  \multicolumn{2}{c|}{3.18} &
  \multicolumn{2}{c|}{8.72} &
  \multicolumn{2}{c|}{15.33} &
  \multicolumn{2}{c|}{1.37} &
  \multicolumn{2}{c}{100} \\ \hline
\multirow{5}{*}{\textbf{L2}}  & 1     & 1e7 & 4.03 & 4.04  & 9.05  & 9.23  &  14.95    & 15.21  & 24.01 & 13.77  & 98.12 &  99.02 \\ \cline{2-13} 
                              & 10    & 1e7 &  4.16    & 4.23  & 9.31&  9.01 & 15.01 &  15.03 & 33.45 &  24.63 & 98.51 &  92.03 \\ \cline{2-13} 
                              & 100   & 1e7 &  4.53    &  5.03 &    9.48  &  14.83 &  14.95 &  18.28 & 48.41 &  32.49 & 94.35 & 95.71  \\ \cline{2-13} 
                              & 1000  & 1e7 &  4.71    &  4.62 &   10.73   & 14.45  &  15.19     &  19.74 & 55.28 &  43.24 & 90.04 & 89.43  \\ \cline{2-13} 
                              & 10000 & 1e7 &  5.33    &  5.59 &   12.74   &  14.15 &   16.26    &  19.26 & 55.83 &  44.04 & 83.44 & 78.25  \\ \toprule[1.5pt]
\textbf{Zheng et al.~\cite{zheng2021using}} &
  1 &
  1e7 &
 3.19  &
 3.18  &
  8.83 &
  8.79 &
  14.09 &
  15.44 &
  30.83 &
  18.12 &
  97.25 & 98.44
   \\ \toprule[0.5pt]
\textbf{Isolated Words} & 1 & 1e7 &   4.22   &  5.47 &  10.15    &  11.19 &   14.31    & 16.24  & 38.07 & 27.62  & 95.14 & 94.33  \\ \toprule[0.5pt]
\multirow{4}{*}{\textbf{EWC}} & 10    & 1e7 &   3.22   &  3.31 &  8.94    &  8.83 &   13.96    & 15.22  & 49.10 & 31.43  & 94.23 & 94.57  \\ \cline{2-13} 
                              & 100   & 1e7 &    3.30  &  3.43 &    8.97  &  9.12 &  13.47      & 15.17  & 59.74 &  45.81 & 90.45 & 89.14  \\ \cline{2-13} 
                              & 1000  & 1e7 &    3.55  &  3.62 &    9.99  &  10.45 &   14.24    & 15.54  & 59.76 &  46.03 & 80.61 & 81.09  \\ \cline{2-13} 
                              & 10000 & 1e7 &   4.01   &  5.37 &    10.28  &  11.23 &   14.15    & 15.87  & 62.19 &  46.71 & 77.17 & 69.42\\
\bottomrule[1.5pt]
\end{tabular}} 
\end{table}

\subsection{Discussion}
New vocabulary emerges all the time due to the evolution of human language. Therefore, it is important to enable a trained ASR system to dynamically acquire unseen vocabulary. The combined loss rescaling and weight consolidation methods proposed in this paper can support continual learning~\cite{parisi2019continual} of an ASR system. The methods neither require any labeled data nor do they require retraining a new ASR model from scratch. 

An interesting finding is that enhancing the gradient of blank tokens within and after OOV words is important as well, which encourages the decoding procedure moving forward, for instance, the rows of $u_{6}$, $u_{8}$ and $u_{10}$ in Figure~\ref{fig:ctc-decoding}. Otherwise, the decoding progress is cut off and models repeatedly produce one token, such as ``news about bre\_ bre\_ bre\_" when only enlarging the gradient of ``bre\_" in the utterance of ``new about bre\_ xi\_ t". Sometimes, the fine-tuned ASR system even repeats one token, for example ``bre\_ bre\_ bre\_ bre\_ bre\_", when $\mu$ is very large. 


Additionally, we find that the performance and the speed of convergence are affected by the batch size, especially for loss rescaling at sentence level. When the batch size is small, e.g.\ 5, all utterances in one batch may contain OOV words, which leads to a bigger rescaled loss. Consequently, the model suffers from gradient explosion, and L2 or EWC regularization can hardly constrain the model weights diverging. 

 \section{Conclusion}
 
 In this paper, we present the use of synthetic speech to boost an ASR model on the recognition of OOV words. In addition to fine-tuning with audio containing OOV words, we propose to rescale loss at sentence level or word level, which encourages models to pay more attention to unknown words. Experimental results reveal that fine-tuning the baseline ASR model combined with loss rescaling and L2/EWC regularization can significantly improve the OOV word recall rate and efficiently overcome models suffering from catastrophic forgetting. Furthermore, loss rescaling at word level is more stable than at sentence level and results in less ASR performance loss on general non-OOV words and previous LibriSpeech tasks. The combination of proposed loss rescaling, which updates the new task-related parameters (OOV word recognition), and EWC, which retains the old task-learned weights (speech recognition on the LibriSpeech dataset), can enable continual learning of an ASR system.
 
 The proposed target word loss rescaling method is simple and effective, but there are still some issues left to be improved. Currently, results are evaluated on synthetic audio data which is different from the spontaneous speech recorded in the real world. Future work could focus on real-scenario speech collecting and model evaluation, which enables us to well understand and compare the contribution of using synthetic and real speech containing OOV words. Additionally, the current OOV word set needs to be known, how to automatically detect and optimize OOV words is a potential direction. The trade-off between WER on universal test sets (e.g.\ LibriSpeech test-clean and test-other sets) and recall rate on the OOV set is another issue. A dynamic L2/EWC weight~\cite{leang2020dynamic} can be adopted to replace the fixed $\lambda$ weight. Later in the fine-tuning, a fixed regularization weight could influence model updating. Moreover, we are interested in investigating the effectiveness of our proposed method on RNN-T and attention-based encoder-decoder ASR systems. It is also worthwhile to explore our loss rescaling method on some general unbalanced label problems, for example, speaker diarization and voice verification. Since continual learning in sequence processing is a young research field~\cite{cossu2021continual, ehret2020continual, ahrens2021drill}, our loss rescaling method may have wider implication for data where novel elements are learnt in temporal or spatial context with known elements.

\section*{Acknowledgments}
The authors gratefully acknowledge the support from the China Scholarship Council (CSC), from the Young Scientists Foundation of Zhejiang Lab, from the German Research Foundation DFG (projects TRR 169 and LeCAREbot) and from the BMWK (projects KI-SIGS and SiDiMo).

\bibliography{main}

\newpage
\section{Appendix}

 \begin{table}[th]
  \setlength{\abovecaptionskip}{0.5cm}
  \setlength{\belowcaptionskip}{0.4cm}
  \caption{The 100 OOV words selected from LRS3 dataset.}
  \label{tab:100-oov-words}
  \centering
  \resizebox{12cm}{!}{
\begin{tabular}{ccccc}
MEANINGFUL     & QUANTUM     & RESILIENCE    & VACCINE          & MONITORING   \\
RESEARCHER     & GLOBALLY    & GLOBALIZATION & EMPOWERMENT      & SPACECRAFT   \\
HORMONE        & HEALTHCARE  & PRESCHOOL     & WORKFORCE        & ALGORITHM    \\
ISRAELI        & NOBEL       & EMPATHY       & ECOSYSTEM        & INTERACTING  \\
SCHIZOPHRENIA  & SOFTWARE    & INTEGRATE     & PROGRAMMED       & YOGA         \\
ALZHEIMER      & VIETNAM     & RETHINK       & PAKISTAN         & LATVIA       \\
PARKINSON      & SOCCER      & RACISM        & SILICON          & MARIJUANA    \\
NEUROSCIENCE   & GENETICALLY & MAINSTREAM    & DEMOGRAPHIC      & UNEMPLOYMENT \\
PARADIGM       & GENOME      & CREATIVITY    & INSULIN          & PROSTHETIC   \\
TRANSGENDER    & KENYA       & RACIST        & STRESSFUL        & PERSONALIZED \\
RAPED          & VIRAL       & STORYTELLING  & ENTREPRENEURSHIP & ADULTHOOD    \\
MICROSOFT      & STEREOTYPE  & EXPERTISE     & LITERACY         & RAINFOREST   \\
MINDSET        & GOOGLE      & WORKPLACE     & YOUTUBE          & SAUDI        \\
FACEBOOK       & IDEOLOGY    & INTERACTIVE   & AUTISM           & CHEMOTHERAPY \\
SUSTAINABILITY & ACTIVIST    & ROBOTIC       & SOCIETAL         & EBOLA        \\
PRODUCTIVITY   & TEENAGER    & DOPAMINE      & SYNDROME         & VIABLE       \\
TRANSFORMATIVE & CORTEX      & OXYTOCIN      & PARENTING        & JIHAD        \\
RECYCLING      & DIMENSIONAL & TARGETED      & LAPTOP           & NIGERIA      \\
SUSTAINABLE    & PROGRAMMING & SMARTPHONE    & COLLABORATIVE    & VEGAN        \\
EINSTEIN       & RESEARCHING & VACCINE       & WIKIPEDIA        & ANTIMATTER 
\end{tabular}} 
\end{table}
\end{document}